\documentclass{article}

\usepackage{graphicx} 
\usepackage{subfigure} 

\usepackage{natbib}

\usepackage{algorithm}
\usepackage{algorithmic}
\usepackage{amsmath}
\usepackage{amsthm,mathtools}
\usepackage{hyperref}
\usepackage{mdwlist}
\usepackage{amssymb}
\usepackage{titling}

\usepackage{paralist}

\def\O{{\mathcal{O}}}

\def\bs1b{\left[s_1\right]}
\def\bs2b{\left[s_2\right]}
\def\bs3b{\left[s_3\right]}

\def\bsa2b{\left[(s_2, a_2)\right]}

\newdimen\pHeight
\newdimen\pLower
\newdimen\pLineWidth
\newdimen\pKern
\newdimen\pIR
\newdimen\pB
\newdimen\pF
\newsavebox{\Cbox}
\newsavebox{\vertCmplx}
\newdimen\Cheight
\newdimen\Cwidth
\sbox{\Cbox}{\rm C} \Cheight=\ht\Cbox \Cwidth=\wd\Cbox
\advance\Cheight by \pHeight
\sbox{\vertCmplx}{\rule[\pLower]{\pLineWidth}{\Cheight}}
\sbox{\Cbox}{\usebox{\Cbox}\kern\pKern\usebox{\vertCmplx}}
\wd\Cbox=\Cwidth

\def\real{{\rm I\kern\pIR R}}

\author{Peeyush Kumar \\
agaron@uw.edu\\
University of Washington Seattle
\and
Doina Precup\\
dprecup@cs.mcgill.ca\\
McGill University, Montreal}

\title{Multi-Timescale, Gradient Descent, Temporal Difference Learning with Linear Options}

\begin{document} 

\begin{titlingpage}

\maketitle
\vskip 0.3in

\begin{abstract} 
Deliberating on large or continuous state spaces have been long standing challenges in reinforcement learning. Temporal Abstraction have somewhat made this possible, but efficiently planing using temporal abstraction still remains an issue. Moreover using spatial abstractions to learn policies for various situations at once while using temporal abstraction models is an open problem. We propose here an efficient algorithm which is convergent under linear function approximation while planning using temporally abstract actions. We show how this algorithm can be used along with randomly generated option models over multiple time scales to plan agents which need to act real time. Using these randomly generated option models over multiple time scales are shown to reduce number of decision epochs required to solve the given task, hence effectively reducing the time needed for deliberation.
\end{abstract} 
\end{titlingpage}

\section{Introduction}
Decision making involves choosing among different actions over a multiple range of time scales. The use of higher level temporally extended actions allow an AI agent to solve an entire task in smaller number of decision epochs. Like many other AI systems, Temporal abstraction is often used in reinforcement learning to solve large problems efficiently. Various methods have been proposed to reduce computational burden, and reducing data complexity in learning, by allowing planning in AI systems to compute a good policy for many situations at once. In environments with continuously valued observations or others with a large number of states, some form of generalization is necessary. Linear methods are perhaps the most common and best understood class of generalization mechanisms. Linear methods encompass a wide spectrum of approaches including look-up tables, state aggregation methods, radial basis functions with fixed bases, etc. Frameworks for temporal abstractions do not usually contain an intrinsic structure to describe such generalizations. To achieve generalizations usually temporal abstractions are defined over parametric representation of the environment states, among which linear representations are the most dominant approach.\par

Since temporal effects are linked to the agents decision method, temporal abstraction framework is tightly coupled with controls. Markov Decision Process (MDP) models provide a formal method of planning in controlled dynamical systems (\cite{sutton:98}). The MDP models for planning which are based on primitive actions can be easily extended to SMDP (Semi Markov Decision Process) planning models which include temporally extended actions. The SMDP theory with models based on \textit{options} framework, provides a formal method for representing temporally abstract information. An option $\mathcal{O}$ is a tuple $\{\mathcal{I}, \mu, \beta\}$, where $\mathcal{I}$ is the initiation set of the option, $\mu$ is the behavior policy of the option $\mathcal{O}$ , $\beta$ is the termination condition which defines the time scale of the option. The option models ($\mu$ and $\beta$)  are either constructed by learning maximum-likelihood primitive models and computing exactly option models by composing primitive models or learning option models using intra-option learning, importance sampling, or eligibility traces. \par

Although the option models have been proved to be quite successful in continuous state spaces (one of the central advantages of temporal abstraction), the ability to plan using temporal abstraction methods is still a challenge when the number of states is large.  This is in large partly because learning option models for large state spaces is very expensive.  Moreover, if the system needs to act in real time, there is not much time for deliberation. In such cases the system needs to make use of many option models over multiple time scales. We propose here that instead of learning option models from the data, the system should rather make use of many option models over multiple time scales even if they are a lot in number. We use the linear option models (\cite{Sorg:2010:LO:1838206.1838211}) for learning and planning. Learning a policy over many option models does not effect the time complexity because the linear methods scale linearly with the complexity of the value function. Linear option model extends the option framework from a tabular representation to  a linear representation.  These models provide a general mechanism for providing temporal abstraction over parametric representation of the state space. Behavior of such a model is proved to be sensible while learning and planning. Although linear options provide great benefits towards time efficiency, the SMDP-LSTD algorithm used for policy evaluation using linear options has a quadratic complexity. This makes their convergence slow as compared to the conventional linear complexity of the TD algorithms.\par

Recently various temporal difference algorithms were proposed which are compatible with both linear function approximation and off-policy training using primitive action in MDP setting (GTD algorithm \cite{Sutton_aconvergent}, GTD2 and TDC algorithms \cite{Sutton09fastgradient-descent}).  These algorithms were shown to converge to the TD fixed point with probability one. We propose a direct extension of the TDC algorithm to the SMDP setup. We show the convergence conditions for the SMDP-TDC algorithm using  results from stochastic approximation theory. \par


\section{Background}\label{sec:bgnd}
An MDP is a tuple $(S, A, P, R, \gamma)$ consisting of a state set $S$, action set $A$, transition function $P : S \times A \times S \mapsto [0, 1]$, an expected reward function $R : S \times A \mapsto \real$, and a discount factor $\gamma \in [0, 1)$. Every MDP time step t, the agent takes an action $a_t$ in state $s_t$, and the environment responds with a reward $r_t$ and resulting next state $s'_t = s_{t+1}$. A policy $\pi : S \times \mapsto [0,1]$ is a mapping from states to actions. An action-value function, $Q(s, a)$ is the expected discounted return obtained by taking action $a$ in state $s$ and following $\pi$ thereafter. The Bellman equation defining the optimal action value function $Q^*$ is:
$$
 Q^*(s,a) = R(s,a) + \gamma \sum_{s' \in S}P^{ss'}_a\max_{b \in A}Q^*(s',a)
$$
A policy that is greedy with respect to a given action value function maps each state to $\arg\max_a Q(s, a)$. The value function $V(s)$ corresponding to a given action value function is the value of the greedy action from each state: $V^\pi (s) = \max_{a} Q^\pi(s, a)$. A policy that is greedy with respect to the optimal value function is the optimal policy. \par


%

\subsection{Off Policy Temporal Difference Learning}
Off Policy methods are an important classes of algorithms studied in Reinforcement Learning. Any controlled dynamical system is said to be off policy, if the behavior policy (the one which is being followed) is different from the policy being evaluated. Among various methods of learning in RL, temporal difference methods are undoubtedly the most important and novel ones. Temporal Difference methods have the advantage that they can learn directly from the raw data without any requirement of the environment's model as well as they can bootstrap by updating estimates based in part on other learned estimates. Among the temporal difference (TD) methods, off policy TD control algorithm known as Q-learning is the most central one. Its simplest form, one-step Q-learning, is defined by
$$
Q(s_t,a_t) =  Q(s_t,a_t) + \alpha\left[r_{t+1} + \gamma\max_{a}Q(s_{t+1}, a) - Q(s_t,a_t)\right]
$$

\subsection{Linear Methods for MDPs}
In reinforcement learning, linear methods have long been used for value function approximation. Instead of using tabular representation of states, states are represented in terms of n dimensional feature vectors $\phi(s) \in \real^n$. Linear methods approximate the value function for a policy $\pi$ as linear combination of state feature vectors. $V^{\pi}(s) \approx \phi(s)^T\theta^{\pi}$, where $\theta^{\pi}$ is the vector parameter to be learned, $\theta \in \real^n$. \par   

Unfortunately Q-learning does not show a good behavior when used with function approximations that are linear in the learned parameters. As a result convergence cannot be guaranteed for such methods (\cite{Baird95residualalgorithms}). Several non-gradient-descent approaches to this problem have been developed, but none has been completely satisfactory . Second-order methods, such as LSTD (\cite{Bradtke96linearleastsquares}, \cite{Boyan02technicalupdate}), can be guaranteed stable under general conditions, but their computational complexity is quadratic in the size of the state space. In recent years several gradient descent methods have been proposed (GTD, GTD2, and TDC algorithms ). These methods are shown to have a linear computation complexity and linear memory requirements. Among these methods TDC shows the fastest rate of convergence. The update estimates for TDC are:

\begin{equation}\label{eq:tdc1}
\theta_{k+1} = \theta_k + \alpha_k(\delta_k\phi_k - \gamma\phi'_k)(\phi^T_k w_k)
\end{equation}

where $\alpha_k$ and $\beta_k$ are the learning rate parameters, and $w_k$ is updated as,

\begin{equation}\label{eq:tdc2}
w_{k+1} = w_k + \beta_k(\delta_k - \phi_k^T w_k)\phi_k
\end{equation}

and $\delta_k$ is the TD error

\begin{equation}\label{eq:tdc3}
\delta_k = r_k + \gamma\theta_k^T\phi_k' - \theta_k^T\phi_k
\end{equation} 

It is proved that TDC algorithm converges to the TD fixed point with probability one.

\subsection{Options and SMDP}
Options are temporally extended actions. They can be thought of as fixed policies that can be invoked in certain set of states and which terminate according to a termination condition. Formally an option is a triple $(\mathcal{I}, \mu, \beta)$, where the set $\mathcal{I} \subset \mathcal{S}$ is the initiation set of the option, $\mu : S \mapsto A$ is the option policy and $\beta : S \mapsto [0,1]$ maps each state to the set $[0, 1]$ which is the probability of termination in each state $s \in S$. MDP along with the options describe a Semi Markov Decision Process (SMDP). SMDP is like a MDP, except the options take varying amounts of time. In each SMDP time step, the agent selects an option and follows the option's policy $\mu$ until termination. In any state, one has to choose an option from the set of options available in that state. When the system enters a new subset of the state space, a new set of options becomes available.\par

The Bellman equation for SMDP setup is a direct extension of the Bellman equation for the MDP case

$$
Q^*_\O(s,o) = r_s^o + \sum_{s'}P^{ss'}_o \max_{o' \in \O '_s}Q^*_\O(s',o') 
$$

where $r_s^o$ is the expected discounted reward  obtained during the execution of the option

$$
r_s^o = E\left[r_t + \dots + \gamma^{k-1} r_{t+k-1}|  s_t = s, o_t = o\right]
$$

and $P^{ss'}_o$ is the option's transition model

Similarly the SMDP version of one step Q learning using options can be written as follows

$$
Q(s,o) = Q(s,o) + \alpha\left[r + \gamma^k \max_{o' \in \O_{s'}}Q(s',o') - Q(s,o)\right]
$$

where $k$ is the number of time steps elapsed between $s$ and $s'$, r denotes the cumulative discounted reward until the termination of the option.

\subsection{Linear Options}
Linear options is a direct extension of the option framework from a tabular representation to a linear representation. Linear options are defined over states which are represented as $n$ dimensional feature vector. A linear option is a tuple $(\mathcal{I}, \mu, \beta)$, where $\mathcal{I} \in \real^n$ is the initiation set, $\mu : \real^n \mapsto A$ is the option policy, and $\beta : \real^n \mapsto [0, 1]$ is the termination condition. \cite{Sorg:2010:LO:1838206.1838211} proposed the idea of linear options as linear representations of options. They proposed an extension of the MDP-LSTD method for policy evaluation in the primitive action case to include options by defining SMDP-LSTD method for behavioral policy evaluation. The solution given by SMDP-LSTD is
\begin{equation}
0 = \sum_{k}^{t}\phi_k(r_k + \gamma_k\phi'^T_k\theta_k - \phi^ T_k\theta_k)
\end{equation}

\begin{equation}
\theta^\pi = \left[\sum_{k=1}^t\phi_k(\phi_k - \gamma_k\phi_k')^T\right]^{-1}\left[\sum_{k=1}^{t}\phi_kr_k\right]
\end{equation}

 A linear option expectation model of a behavior policy's dynamics is denoted by $(F_\pi, b_\pi)$ which attempt to satisfy, for all time steps t.

\begin{equation}
F_\pi\phi_t \approx E\left[\gamma_t\phi'_t|\phi_t\right]
\end{equation}

\begin{equation}
b_\pi\phi_t \approx E\left[r_t|\phi_t\right]
\end{equation} 

where $F_\pi\phi$ can be interpreted as the expected discounted termination feature vector and $b_\pi\phi$ can be interpreted as the expected discounted reward until termination. A learning and planning agent will need to estimate $F_\pi$ and $b_\pi$ from data. Given an input feature vector $\phi$, the following recursion defines a LOEM policy evaluation update to the value function
parameters $\theta_k$:

\begin{equation}
\theta_{k+1} = \theta_{k} +\alpha_k(b_\pi^T\phi + \theta_k^TF_\pi\phi - \theta_k^T\phi)\phi
\end{equation}

LOEM policy evaluation updates is shown to converge to the same behavior value function parameters as does the SMDP-LSTD algorithm.

\section{Gradient Descent TD algorithm for linear options}\label{sec:tdc}
The LSTD algorithms developed can be guaranteed stable under general conditions but they are quadratic in computational complexity. We propose here a gradient descent method (SMDP-TDC ) for planning in Reinforcement learning using off-policy TD framework. SMDP-TDC is the direct extension of TDC algorithm (Equations~\ref{eq:tdc1}, \ref{eq:tdc2} and \ref{eq:tdc3}) to the SMDP setting.  The option value function is approximated as $V^{\pi}(s) = \max_o Q^\pi(s,o)  \approx \theta^T\phi_s$.  We aim to minimize the mean-square projected bellman error ($MSPBE_\theta = \left\| V_\theta - \Pi TV_\theta\right\|^2_D$. For the linear option architecture $V_\theta = \Phi\theta$ where $\Phi$ is the matrix whose rows are $\phi_s$. $\Pi$ is the projection operator, $\Pi = \Phi(\Phi^T D \phi)^{-1}\Phi^T D$ (where D is the diagonal matrix containing stationary distribution), and $T$ is the SMDP Bellman Operator. 
\begin{flalign*}
&MSPBE_\theta & \\
&= \left\| V_\theta - \Pi TV_\theta\right\|^2_D & \\
&=\left\| \Pi(V_\theta - TV_\theta)\right\|^2_D & \\
&=(\Pi(V_\theta - TV_\theta))^TD(\Pi(V_\theta - TV_\theta)) & \\
&=((V_\theta - TV_\theta))^T\Pi^TD\Pi((V_\theta - TV_\theta)) & \\
&= ((V_\theta - TV_\theta))^TD^T \Phi(\Phi^T D \phi)^{-1}\Phi^T D((V_\theta - TV_\theta)) & \\
&= (\Phi^TD(TV_\theta-V_\theta))^T(\Phi^TD\Phi)^{-1}(\Phi^TD(TV_\theta-V_\theta)) & \\
&= E[\delta\phi]^TE[\phi\phi^T]^{-1}E[\delta\phi] &
\end{flalign*}

Note that
\begin{flalign*}
E[\phi\phi^T] &= \sum_s{d_s\phi_s\phi_s^T} = \Phi D\Phi^T \\
E[\delta\phi] &= \sum_s d_s\phi_s\left(r + \gamma^l\sum_{s'}P_{ss'}^oV_\theta(s')-V_\theta(s)\right)\\
&= \Phi^TD(TV_\theta-V_\theta)
\end{flalign*}

Therefore,

\begin{flalign*}
& -\frac{1}{2}\nabla MSPBE_\theta & \\
&= E[(\phi-\gamma^l\phi')\phi^T]E[\phi\phi^T]^{-1}E[\delta\phi] & \\
&=(E[\phi]-E[\gamma^l\phi'\phi^T])E[\phi\phi^T]^{-1}E[\delta\phi] & \\
&=E[\delta\phi]-\gamma^lE[\phi'\phi^T]E[\phi\phi^T]^{-1}E[\delta\phi] & \\
&=E[\delta\phi]-\gamma^lE[\phi'\phi^T]w &
\end{flalign*}

where $w \approx E[\phi\phi^T]^{-1}E[\delta\phi]$. Using this form for the gradient of MSPBE, we can write down the gradient descent expression (parameterized by $\alpha$ and $\beta$) to minimize MSPBE as follows:

\begin{equation}\label{eq:smdptdc1}
\theta_{k+1} = \theta_k + \alpha_k(\delta_k\phi_k - \gamma^l\phi'_k)(\phi^T_k w_k)
\end{equation}

where $w_k$ is updated as,

\begin{equation}\label{eq:smdptdc2}
w_{k+1} = w_k + \beta_k(\delta_k - \phi_k^T w_k)\phi_k
\end{equation}

and $\delta_k$ is the TD error

\begin{equation}\label{eq:smdptdc3}
\delta_k = r_k + \gamma^l\theta_k^T\phi_k' - \theta_k^T\phi_k
\end{equation} 

where $l$ is the length of the option being executed.

Convergence proof of SMDP-TDC can be carried out in a similar way as in \cite{Sutton09fastgradient-descent}. Hence it can be shown that SMDP-TDC converges to the TD fixed point with probability one. 

\subsection{Convergence Proof for SMDP-TDC} 
\textbf{Theorem 1} \textit{Consider the iterations \ref{eq:smdptdc1} and \ref{eq:smdptdc2} of the SMDP-TDC algorithm. Let the step-size sequences $\alpha_k$ and $\beta_k$, $k\geq0$ satisfy in this case $\alpha_k$, $\beta_k > 0$, for all $k$, $\sum_{k=0}^{\infty}\alpha_k = \sum_{k=0}^\infty\beta_k = \infty$, $\sum_{k=0}^\infty\alpha_k^2, \sum_{k=0}^\infty\beta_k^2 < \infty$ and that $\frac{\alpha_k}{\beta_k} \rightarrow 0$ as $k \rightarrow \infty$. Further assume that $(\phi_k, r_k, \phi'_k)$ is an i.i.d sequence with uniformly bounded second moments. Let $A = E[\phi_k(\phi_k - \gamma^l\phi_k')^T],\ b = E[r_k\phi_k]$, and $C = E[\phi_k\phi_k^T]$. Assume that A and C are non singular. Then the parameter vector $\theta_k$ converges with probability one to the TD fixed point.}\\
{\large\textit{Proof}} The proof of this theorem is based on a two time scale difference in the step-size schedule $\{\alpha_k\}$ and $\{\beta_k\}$; refer \cite{Borkar1997291} for a convergence analysis of the general two timescale stochastic approximation recursions. The recursions \ref{eq:smdptdc1} and \ref{eq:smdptdc2} correspond to the faster and slower recursions respectively. This is because beyond some integer $N_0 >0\ \forall \ k$, the increments in \ref{eq:smdptdc1} are uniformly larger than those in \ref{eq:smdptdc2} and hence converge faster. Along faster timescale,  i.e, the one corresponding to $\{\beta_k\}$, the associated system ODEs corresponds to

\begin{equation}\label{eq:1}
\dot{\theta}(t) = 0
\end{equation}

\begin{equation}\label{eq:2}
\dot{w}(t) = E[\delta_t\phi_t | \theta(t)] - Cw(t)
\end{equation}

The ODE \ref{eq:1} suggests that $\theta(t) = \theta$ from the faster timescale perspective. Indeed, recursion \ref{eq:smdptdc1} can be rewritten as $\theta_{k+1} = \theta_k + \beta_k\xi(k)$, where from \ref{eq:smdptdc1}, $\xi(k) = \left( \frac{\alpha_k}{\beta_k} (\delta_k\phi_k - \gamma^l \phi'_k\phi_k^Tw_k)\right) \rightarrow 0$ almost surely as $k \rightarrow \infty$ because $\frac{\alpha_k}{\beta_k} \rightarrow 0$ as $k \rightarrow \infty $. By the Hirsch lemma (\cite{Hirsch:1989:CAD:70405.70406}), it follow that $\left\|\theta_k - \theta\right\| \rightarrow 0$ almost surely as $k \rightarrow \infty$ for the same $\theta$ that depends on the initial condition $\theta_0$ of recursion  \ref{eq:smdptdc1}.\par

Consider now the recursion \ref{eq:smdptdc2}. Let $M_{k+1} = (\delta_k\phi_k - \phi_k\phi_k^Tw_k) - E[(\delta_k\phi_k - \phi_k\phi_k^Tw_k)|\mathcal{F}(k)]$, where $\mathcal{F}(k) = \sigma(w_m, \theta_m, m \leq k;\phi_s,\phi_s',r_s, s < k),\ k\geq 1$ are the sigma fields generated by $w_0, \theta_0, w_{m+1}, \theta_{m+1}, \phi_m, \phi'_m, 0\leq m <k$. It is easy to verify that $M_{k+1}, k\geq 0$ are integrable random variables that satisfy $E[M_{k+1}|\mathcal{F}(k)] = 0,\ \forall k \geq 0$. Also because $r_k,\ \phi_k$ and $\phi'_k$ have uniformly bounded second moments, it can be seen that

$$
E \left[ \left\| M_{k+1} \right\|^2 | \mathcal{F}(k)\right] \leq c_1\left(1 + \|| w_k||^2 + || \theta_k || ^2\right),
$$

for all $k \geq 0$, for some constant $c_1 >0$ Now consider the ODE pair \ref{eq:1} and \ref{eq:2}. Because $\theta(t) = \theta$ from \ref{eq:1}, the ODE \ref{eq:2} can written as

\begin{equation}\label{eq:3}
\dot{w}(t) = E[\delta_t\phi_t|\theta] - C w(t)
\end{equation}

Now consider the function $h(w) = E[\delta\phi | \theta] - Cw$, i.e. the driving vector field of the ODE \ref{eq:3}. For \ref{eq:3}, $w^* = C^{-1} E[\delta\phi|\theta]$ is the unique globally asymptotically stable equilibrium. Let $h_{\infty}(.)$ be the funciton defined by $h_\infty(w) = \lim_{r \rightarrow\infty} \frac{h(rw)}{r}$. Then $h_\infty(w) = -Cw$. For the ODE $ \dot{w}(t) = h_\infty(w(t)) = -Cw(t)$ the origin is a globally asymptotically stable equilibrium because C is a positive definite matrix. The assumptions are now verified and by \cite{Borkar00theode} Theorem 2.2 we obtain $||w_k - w^*|| \rightarrow 0$ almost surely as $k \rightarrow \infty$.\par

Consider now the slower time scale recursion \ref{eq:smdptdc1}. Using above, \ref{eq:smdptdc2} can be rewritten as
\begin{equation}\label{eq:5}
\theta_{k+1} = \theta_k + \alpha_k(\delta_k\phi_k - \gamma^l\phi'_k\phi_k^TC^{-1}E[\delta_k\phi_k|\theta_k])
\end{equation}   

Let $\mathcal{G}(k) = \sigma(\theta_m, m \leq k; \phi_s, \phi_s', r_s, s<k)$ be the sigma fields generated by the quantities $\theta_0, \theta_{m+1}, \phi_m, \phi_m',0 \leq m < k$. Let

\begin{flalign*}
&Z_{k+1} = (\delta_k\phi_k - \gamma^l\phi'_k\phi_kC^{-1}E[\delta_k\phi_k|\theta_k])&\\
&-E[(\delta_k\phi_k - \gamma^l\phi_k'\phi_k^TC^{-1}E[\delta_k\phi_k \ \theta_k])|\mathcal{G}(k)]&\\
&= (\delta_k\phi_k - \gamma^l\phi_k'\phi_k^TC^{-1}E[\delta_k\phi_k | \theta_k])&\\
& -E[\delta_k\phi_k|\theta_k] -\gamma^lE[\phi_k'\phi_k^T]C^{-1}E[\delta_k\phi_k | \theta_k]
\end{flalign*}

It is easy to see that $Z_k, k\geq0$ are integrable random variables and $E\left[Z_{k+1}|\mathcal{G}(k)\right] = 0, \forall k \geq0$. Further,
$$
E\left[\left\|Z_{k+1}\right\|^2|\mathcal{G}(k)\right] \leq c_2\left(1 + \left\|\theta_k\right\|^2\right), k \geq 0
$$
for some constant $c_2 > 0$ again because $r_k, \phi_k$ and $\phi_k'$ have uniformly bounded second moments. \par
Consider now the following ODE associated with \ref{eq:smdptdc1}
\begin{equation}\label{eq:6}
\dot{\theta}(t) = \left(I - E\left[\gamma^l\phi'\phi^t\right]C^{-1}\right)E[\delta\phi|\theta(t)]
\end{equation}
 
Let $\bar{h}(\theta(t))$ be the driving vector field of the ODE \ref{eq:6}. Note that 
\begin{align*}
\bar{h}(\theta(t)) &= \left(I - E\left[\gamma^l\phi'\phi^T\right]C^{-1}\right)E \left[\delta\phi|\theta(t) \right]\\
&=\left( C - E \left[ \gamma^l\phi'\phi^T \right] \right)C^{-1}E \left[ \delta\phi|\theta(t) \right]\\
&=\left( E \left[ \phi\phi^T \right] - E \left[ \gamma^l\phi'\phi^T \right] \right) C^{-1} E \left[ \delta\phi|\theta(t) \right]\\
&= A^TC^{-1}(-A\theta(t) + b)
\end{align*}

 because $E[\delta\phi|\theta(t)] = -A\theta(t) + b$.\par
 
 Now $\theta^* = A^{-1}b$ can be seen to be the unique globally asymptotically stable equilibrium for \ref{eq:6}. Let $\bar{h}_\infty(\theta) = \lim_{r\rightarrow \infty} \frac{\bar{h}(r\theta)}{r}$. Then $\bar{h}_\infty(\theta) = -A^TC^{-1}A\theta$. Consider now the ODE
\begin{equation}\label{eq:7}
\dot{\theta}(t) = -A^TC^{-1}A\theta(t)
\end{equation}

Because $C^{-1}$ is positive definite and $A$ has full rank (as it is nonsingular by assumption), the matrix $A^TC^{-1}A$ is also positive definite. The ODE \ref{eq:7} has the origin as its unique globally asymptotically stable equilibrium. The assumptions are once again verified and the claim follows.\hfill

\section{Multiple Time Scales and Real Time RL}\label{sec:multi}
Autonomous systems make decisions by choosing actions at multiple time scales. Although planning for acting at multiple time scales requires more computational time. Real time systems do not have much time for deliberation. Hence a lot of research has been done to construct just the right number of options by optimizing the computational time around models  to accommodate multiple time scales (\cite{Stolle02learningoptions}, \cite{Wolfe05identifyinguseful}). Is it necessary to  be so careful, or can we do around with many option models? The ideas central to planning is searching and learning while searching. We demonstrate empirically that it is much better to search with many option models, rather than using few handcrafted ones. \par
\begin{figure}[htpb]     
\centering       
  \includegraphics[width=0.5\textwidth]{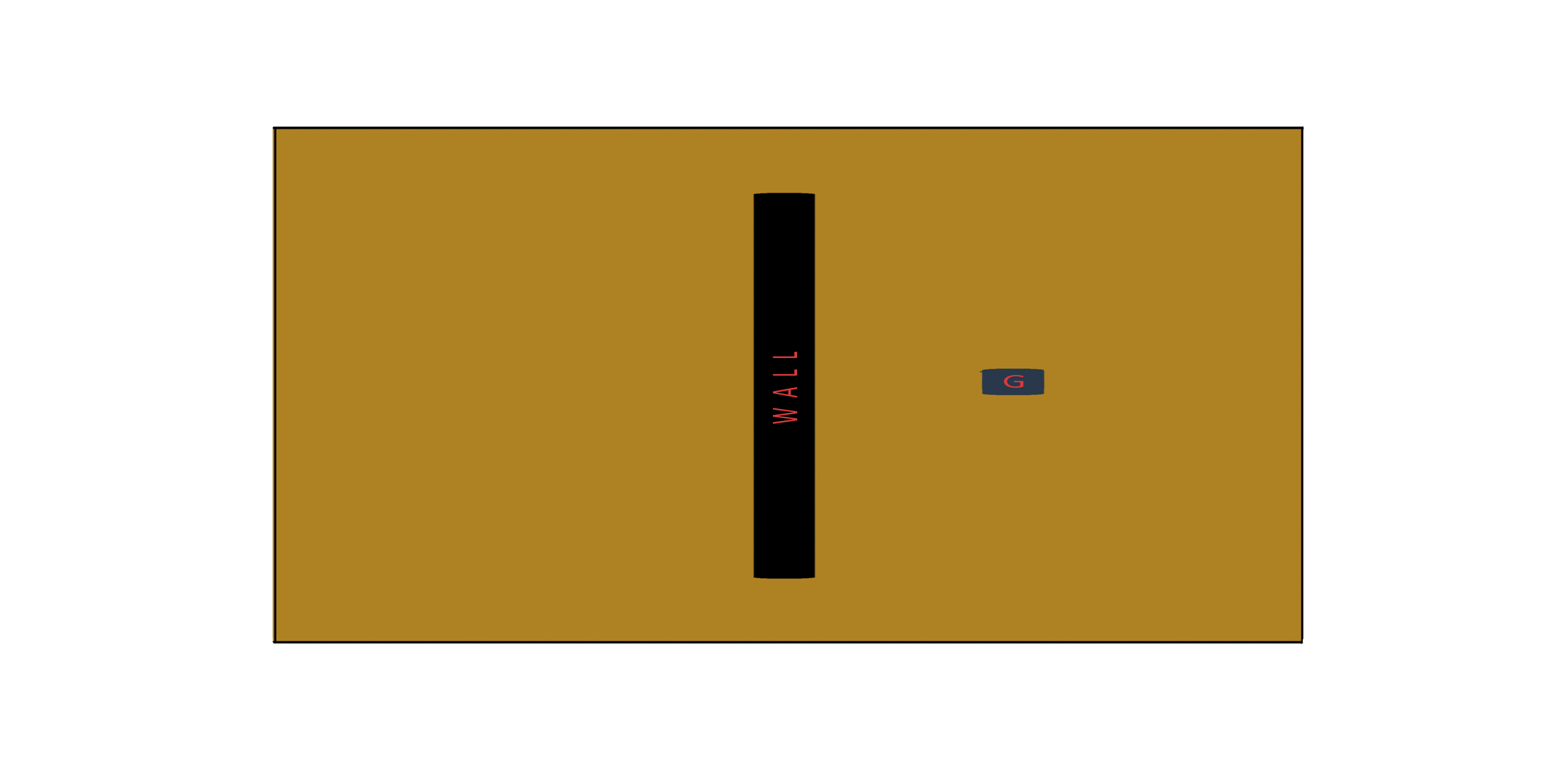}
  \caption{Grid World Domain}
  \label{fig:gridworld}
\end{figure} 

We use the sample based search to search through the state space. The search is done by picking up a quasi-random model and sampling the next state, up to a given depth. At the given depth a guess of the value function is used to greedily pick up the best move. Typically a large number of rollouts is performed, and values lower in the search tree are used to update the values higher up in the search tree (\cite{Kocsis06banditbased}, \cite{veness:reinforcement}). Number of rollouts is a parameter that controls the time for deliberation. The experiment was performed on the gridworld domain shown in Figure~\ref{fig:gridworld}. We use the \textit{hallway going options} where the option models are precoded. Values for all options are learned over 50 trials, using off-policy random behavior. The following six experimental conditions were used to describe what to do on the next time step  \begin{inparaenum}[\itshape a\upshape)]
\item Execute the greedy policy using only primitives \item Pick the primitive action suggested by the greedy option policy with
the hand-crafted options \item Pick the primitive action suggested by the greedy policy with random options \item Do 100 rollouts of depth 3 using only primitive actions \item Do 100 rollouts of depth 3 using primitive actions and crafted options \item Do 100 rollouts of depth 3, sampling among 6 options from the random set
\end{inparaenum} 

\begin{figure}[htpb]     
\centering       
  \includegraphics[width=0.4\textwidth]{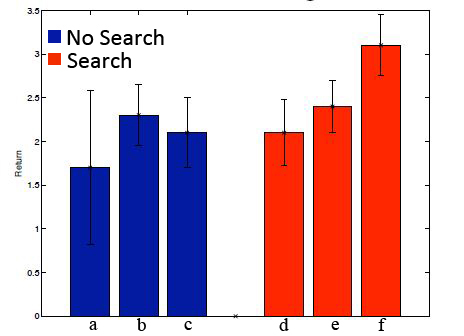}
  \caption{Returns for search}
  \label{fig:search}
\end{figure} 

As wee see in Figure~\ref{fig:search} search using many option models is better than just using a few handcrafted ones. The results also demonstrate the usefulness of temporal abstraction and search. Search, and learning using many option model does not effect the time, as the computational complexity of the linear option framework does not depend upon the number of options being used. There are different ways to incorporate search in a real-time system \begin{inparaenum}[\itshape a\upshape)]\item The planning can be separately threaded \item Trying to anticipate what may happen using the model and planning for different states before the sensation is received \end{inparaenum}. \par

\section{Empirical Results}\label{sec:experiments}
To assess the practical utility of the proposed framework we demonstrate here some experiments where the system learns policy over multiple time scale linear option models updating estimates using the SMDP-TDC algorithm. We also demonstrate the time scales on which the system acts depending upon its position in the environment and its immediate neighbors. The results are demonstrated over 2 domains. The first domain is a grid world domain as shown in Figure~\ref{fig:gridworld} and the second domain is the continuous room domain with large dimension feature representation.\par

\subsection{Grid World Domain}
\begin{figure}[htpb]     
\centering       
  \includegraphics[width=0.5\textwidth]{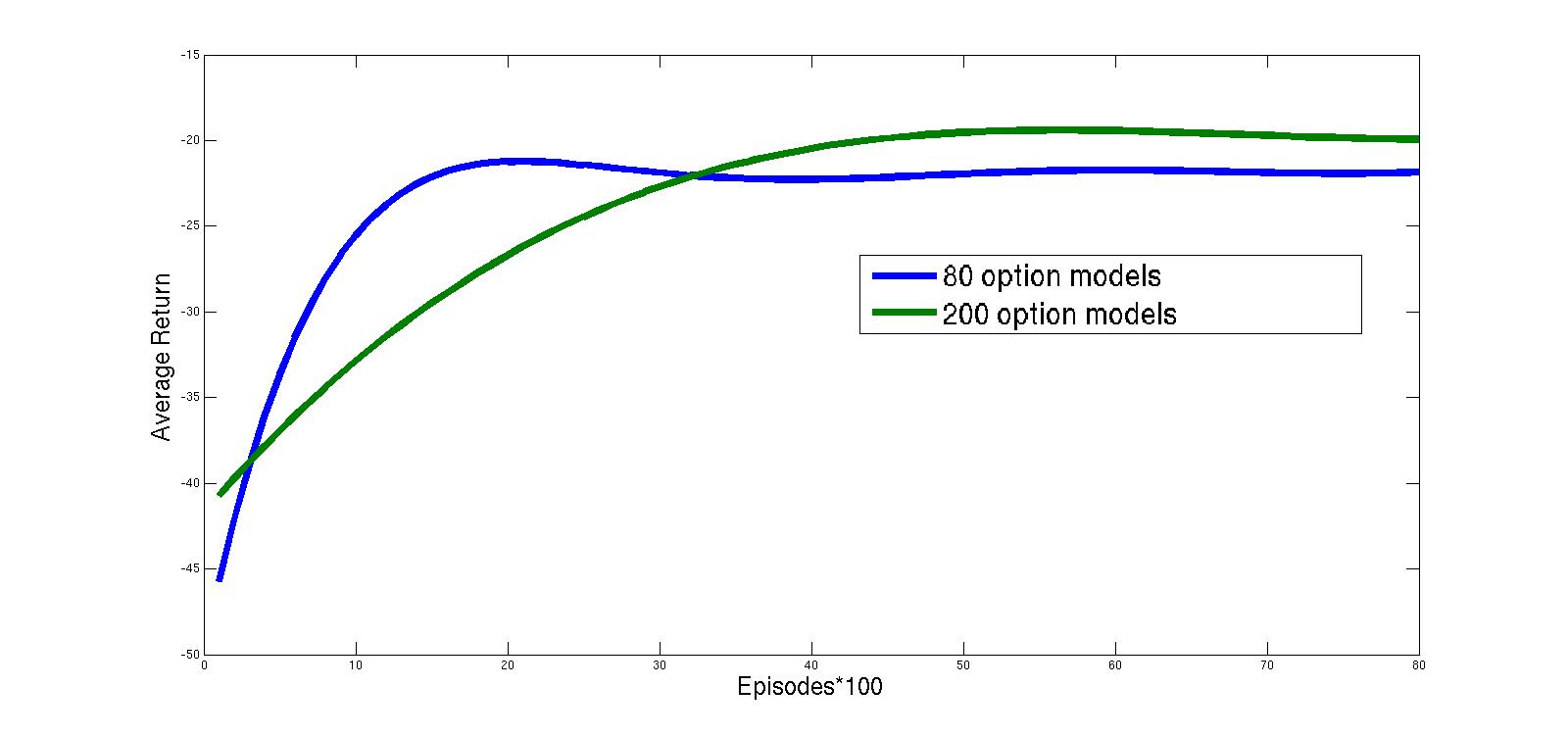}
  \caption{Average Return compared for 2 experimental conditions with options over different multiple time scales (a) 2 timescales of 0.5, 1; and b) 5 time scales of 0.2, 0.4, 0.6, 0.8, 1) by policy learned using SMDP-TDC.}
  \label{fig:gwavg_return}
\end{figure}

The system starts from any state in the state space, it receives a reward of +10 when it reaches the goal state(blue box in Figure~\ref{fig:gridworld}), while it receives a reward of -3 when it hits the wall and a reward of -1 for all the other transitions. The system has four primitive actions, while acting in the environment with 15\% noise. We demonstrate two experimental conditions where option models are composed over multiple time scales and various random policies. The first experiment consists of two time scales - $\beta = 0.5,\ 1$ and biased random policies defined as follows - Pick a probability $x \in (0,1)$, pick uniformly an action $a$, the system will pick the action $a$ with probability $x$ and all other actions uniformly with probability $1-x$. The sytem has same termination conditions, and pick primitive actions with respect to the same policy for all states. Likewise there are 40 policies, hence the system has 80 options available at all states. The second experimental condition consists of 5 termination conditions - $\beta = 0.2, 0.4, 0.5 0.8, 1$ with the same set of policies and hence it has 200 options available at all states. We plot the average return encounterd by the system averaged over various multiple runs in Figure~\ref{fig:gwavg_return}.\par

Important observations which we note in the experiments are that \begin{inparaenum}[\itshape a\upshape)] \item Far away from the goal, the best options are the ones biased towards the goal, and with a larger time scale. \item Close to the wall, the best options are the ones biased towards the goal and with a shorter time scale. \item Still closer, the best options are biased up or down. \item The system which does not decide on each time step sometimes hits the wall. \item It is much better to search with a lot of random options over multiple time scales rather than a few handcrafted ones \end{inparaenum}. \par

\begin{figure}[htpb]     
\centering       
  \includegraphics[width=0.5\textwidth]{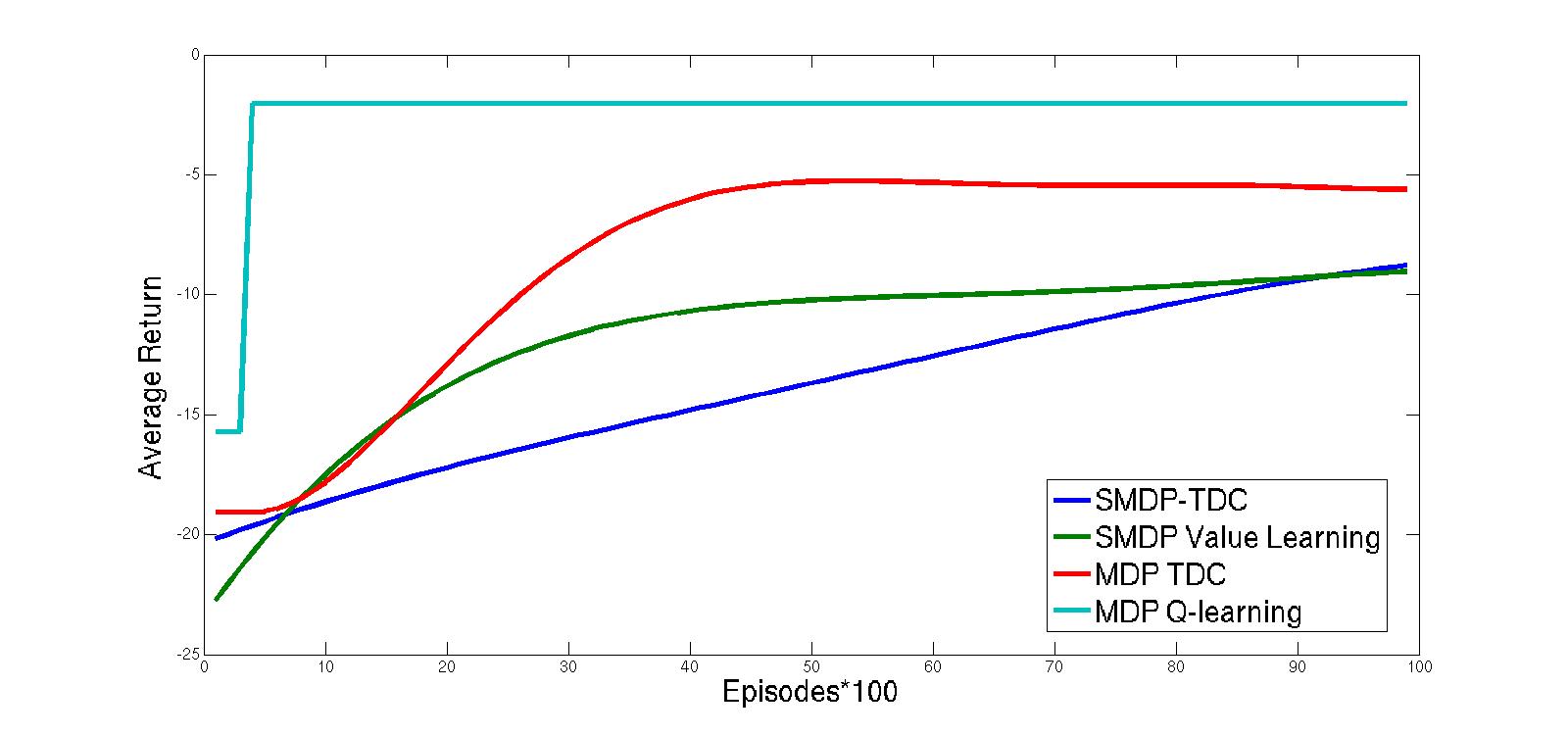}
  \caption{Average Return on the Grid World Domain.}
  \label{fig:gwsmdptdc}
\end{figure} 

In Figure~\ref{fig:gwsmdptdc} we compare the performance of SMDP-TDC method with the primitive TDC method. We also compare the same results without using any function approximation while following a) SMDP Value Learning, and b)MDP Q learning.  We observe the convergence of SMDP-TDC algorithm to a recursive optimal solution. In Figure~\ref{fig:gwsmdptdc} we observe that option models with multiple time scales reach the goal state more often than its primitive counter part. 

\begin{figure}[htpb]     
\centering       
  \includegraphics[width=0.5\textwidth]{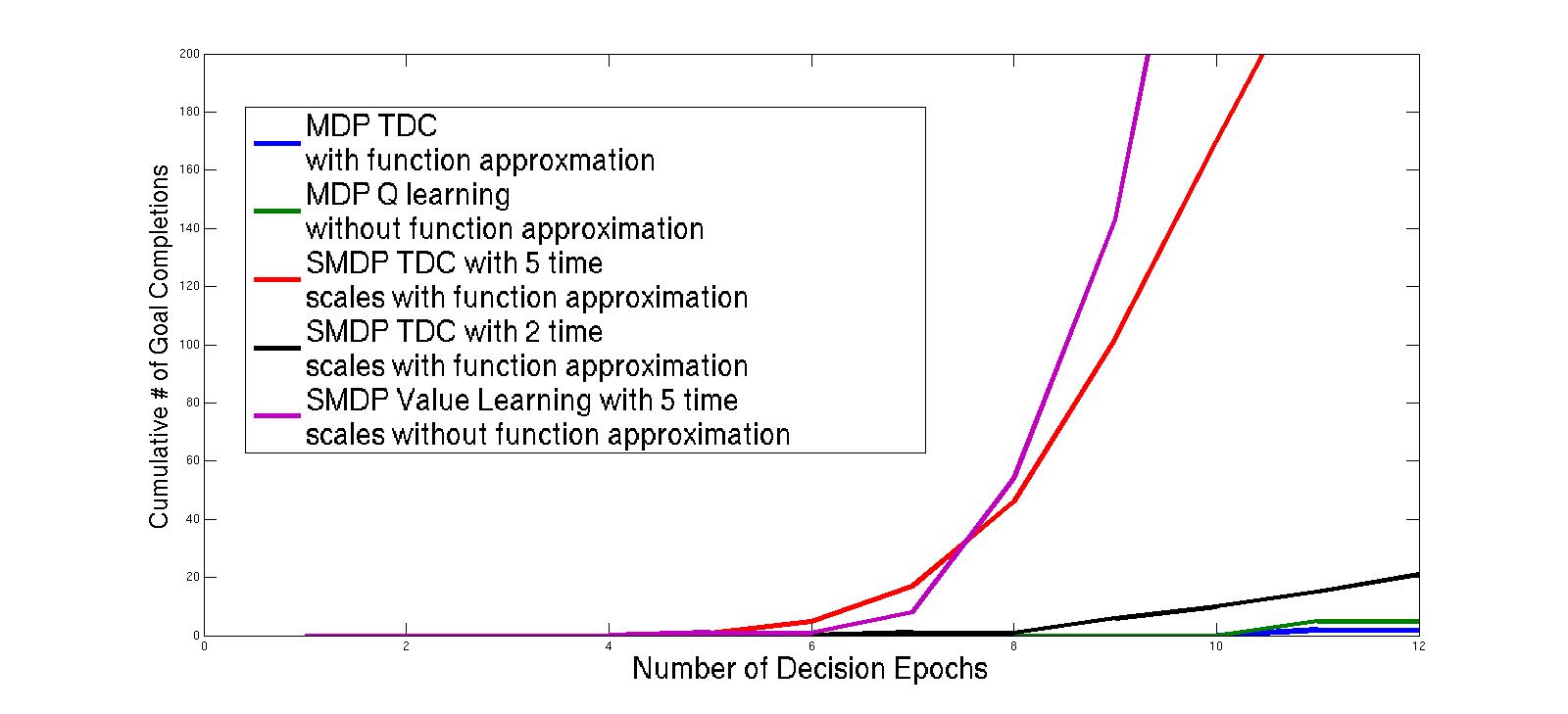}
  \caption{Performance in the Grid World Domain}
  \label{fig:gwgoalcompletions}
\end{figure} 

\subsection{Continuous Domain}
\begin{figure}[htpb]     
\centering       
  \includegraphics[width=0.3\textwidth]{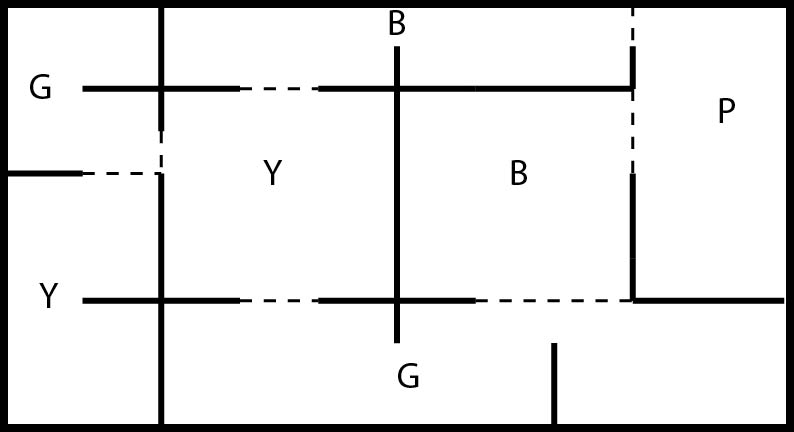}
  \caption{Continuous Room Domain}
  \label{fig:contdomain}
\end{figure}

We ran our second set of experiments on a continuous navigation domain as described in \cite{Sorg:2010:LO:1838206.1838211}. The domain (Figure~\ref{fig:contdomain}), which is $10\times 10$ continuous room world consists of rooms which are separated by walls as shown. In addition the floor of each room is colored with one of the 4 colors (P)urple, (G)reen, (Y)ellow, (B)lue. In figure the floor colors are indicated by the respective first letter, and rooms are seperated by using dashed lines. While the walls are shown using solid lines.\par
The agent controls a circular wheel robot for navigation that is capable of observing its current global position $[x,y] \in [0, 10]^2$ and its current global orientation in terms of the an angle $\psi \in[0,360]$. In addition it can also detect the color of the floor beneath it. It has 3 available primitive actions:\texttt{forward} which moves the robot one unit in the forward direction with respect to its current position and orientation, \texttt{left} turns the robot towards left by an angle of $30^o$ at its current location, and \texttt{right} which turns the robot towards right with an angle of $30^o$ at its current location. If the robot hits the wall, its motion is halted in the direction perpendicular to the wall plane. The agent is given a reward of 1 for reaching the purple room on the right side of the domain. After receiving its reward, the agent is transported to the yellow room on the left corner of the building. At all other times, the agent receives a small negative reward of 0.01.\par
The agent represent states using feature vectors $\phi$ of 1204 features. The first four features encode the color of the floor in a binary format. The rest of the features are calculated by placing radial basis functions every one step in the $x-y$ direction and every $30^o$ for $\psi$. The feature values are calculated using $\phi_i = b exp(- \frac{1}{2}(s - u_i)^TC(s - u_i))$ where $s = [x, y, \psi]$, $b = 10$, $C = diag( \frac{1}{1.2}, \frac{1}{1.2},  \frac{1}{30})$, where $u_i$ are the center of the radial basis functions. To make the feature vector sparse, which can help computationally, any feature that would have had a value $\phi_i < 0.1$, is set to 0. 

\begin{figure}[htpb]     
\centering       
  \includegraphics[width=0.5\textwidth]{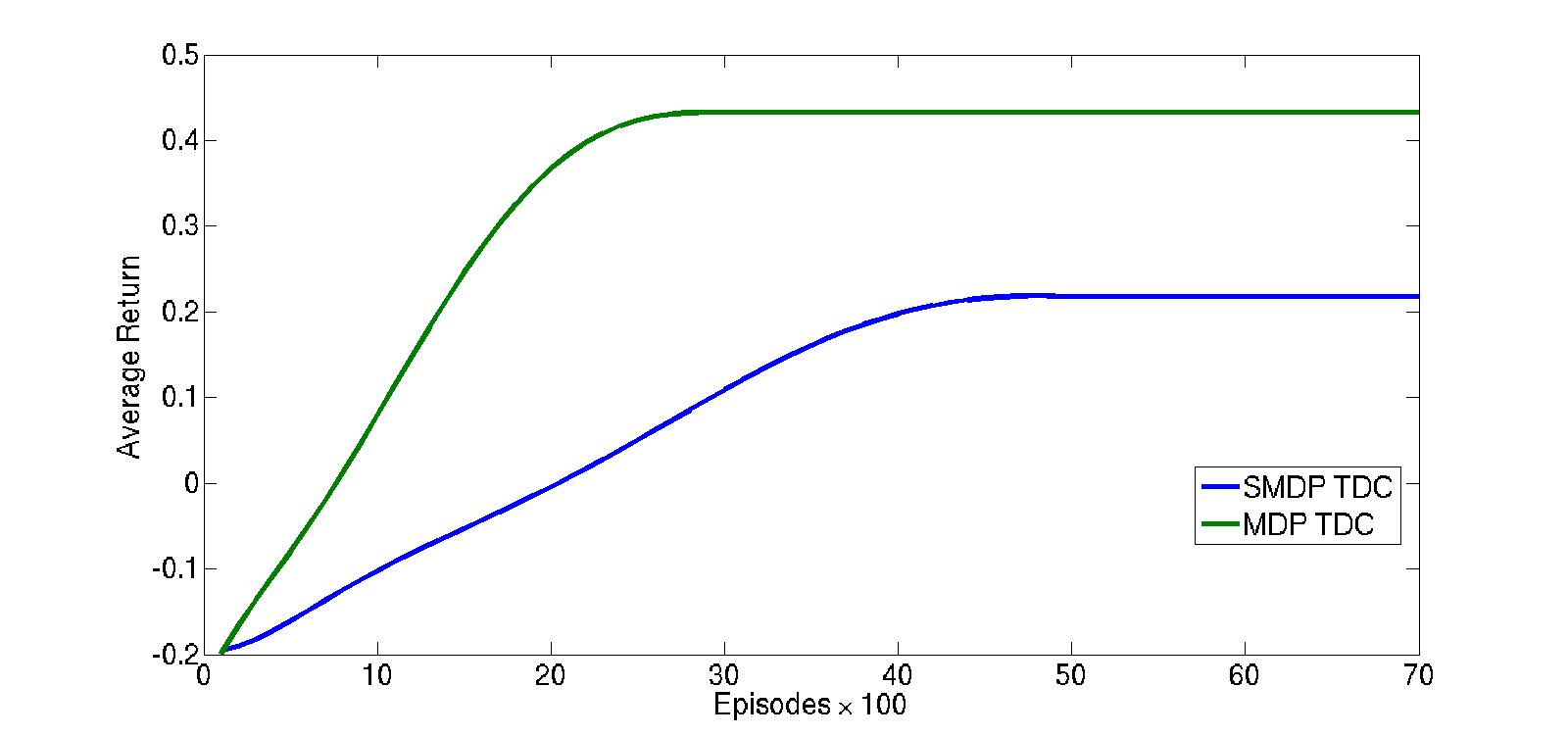}
  \caption{Average Return in the continuous Domain}
  \label{fig:contaverage}
\end{figure}

\begin{figure}[htpb]     
\centering       
  \includegraphics[width=0.5\textwidth]{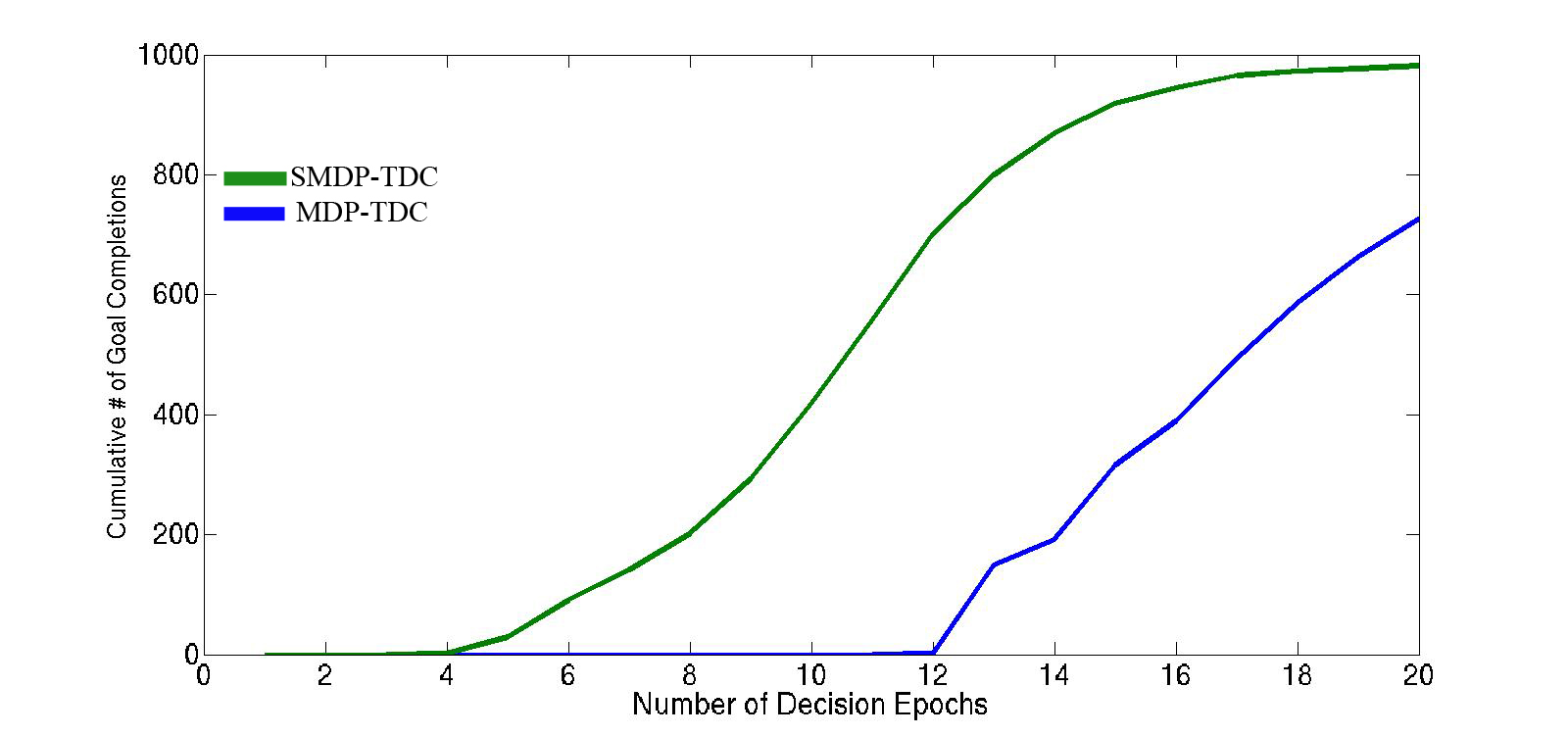}
  \caption{Performance in the continuous Domain}
  \label{fig:contgoalcompletions}
\end{figure}

We tested 2 agents with the above setting, the first agent uses the MDP-TDC learning, while the second agent uses the SMDP-TDC learning using 200 randomly generated option models, with 5 different time scales. The average return obtained is reported in Figure~\ref{fig:contaverage}. While Figure~\ref{fig:contgoalcompletions} shows that the number of goal completions are more when the agent uses option models over multiple time scales as compared to the primitive models. 

\section{Conclusion}
Systems which need to act real time, do not have much time for deliberation. For such systems we propose the use of many option models composed over multiple time scale. We also propose an SMDP learning algorithm which is proved to be convergent with the linear option models while achieving a linear time efficiency and memory requirement. We demonstrate our results over the gridworld domain and a continuous room domain. We observe that systems performing while using an approximate representation of the environment states and learning using SMDP TDC algorithm over multiple time scales performs better in all aspects as compared to other methods. They show a recursive optimal behavior which is reasonably good as compared to the optimal solution. Additionally they achieve goals more efficiently as compared to other methods which deliberate at every time step. It will be interesting to see the extension of SMDP TDC algorithm using multiple time-step TD and eligibility traces.      

\bibliography{example_paper}

\begin{thebibliography}{}

\bibitem[Baird, 1995]{Baird95residualalgorithms}
Baird, L. (1995).
\newblock Residual algorithms: Reinforcement learning with function
  approximation.
\newblock In {\em In Proceedings of the Twelfth International Conference on
  Machine Learning}, pages 30--37. Morgan Kaufmann.

\bibitem[Borkar, 1997]{Borkar1997291}
Borkar, V.~S. (1997).
\newblock Stochastic approximation with two time scales.
\newblock {\em Systems amp; Control Letters}.

\bibitem[Borkar and Meyn, 2000]{Borkar00theode}
Borkar, V.~S. and Meyn, S.~P. (2000).
\newblock The o.d.e. method for convergence of stochastic approximation and
  reinforcement learning.
\newblock {\em SIAM J. CONTROL OPTIM}.

\bibitem[Boyan, 2002]{Boyan02technicalupdate}
Boyan, J.~A. (2002).
\newblock Technical update: Least-squares temporal difference learning.
\newblock In {\em Machine Learning}.

\bibitem[Bradtke et~al., 1996]{Bradtke96linearleastsquares}
Bradtke, S.~J., Barto, A.~G., and Kaelbling, P. (1996).
\newblock Linear least-squares algorithms for temporal difference learning.
\newblock In {\em Machine Learning}, pages 22--33.

\bibitem[Hirsch, 1989]{Hirsch:1989:CAD:70405.70406}
Hirsch, M.~W. (1989).
\newblock Convergent activation dynamics in continuous time networks.
\newblock {\em Neural Netw.}

\bibitem[Kocsis and Szepesvári, 2006]{Kocsis06banditbased}
Kocsis, L. and Szepesvári, C. (2006).
\newblock Bandit based monte-carlo planning.
\newblock In {\em In: ECML-06. Number 4212 in LNCS}.

\bibitem[Sorg and Singh, 2010]{Sorg:2010:LO:1838206.1838211}
Sorg, J. and Singh, S. (2010).
\newblock Linear options.
\newblock AAMAS '10.

\bibitem[Stolle and Precup, 2002]{Stolle02learningoptions}
Stolle, M. and Precup, D. (2002).
\newblock Learning options in reinforcement learning.
\newblock In {\em Lecture Notes in Computer Science}.

\bibitem[Sutton and Barto, 1998]{sutton:98}
Sutton, R. and Barto, A. (1998).
\newblock {\em Reinforcement {Learning}: {An} {Introduction}}.
\newblock MIT Press.

\bibitem[Sutton et~al., 2009]{Sutton09fastgradient-descent}
Sutton, R.~S., Maei, H.~R., Precup, D., Bhatnagar, S., Silver, D., Szepesvári,
  C., and Wiewiora, E. (2009).
\newblock Fast gradient-descent methods for temporal-difference learning with
  linear function approximation.
\newblock In {\em In Proceedings of the 26th International Conference on
  Machine Learning}.

\bibitem[Sutton et~al., 2001]{Sutton_aconvergent}
Sutton, R.~S., Szepesvári, C., and Maei, H.~R. (2001).
\newblock A convergent o(n) algorithm for off-policy temporal-difference
  learning with linear function approximation.
\newblock In {\em Advances in Neural Information Processing Systems 21}. MIT
  Press.

\bibitem[Veness et~al., 2010]{veness:reinforcement}
Veness, J., Ng, K.~S., Hutter, M., and Silver, D. (2010).
\newblock Reinforcement learning via aixi approximation.
\newblock In {\em AAAI'10}.

\bibitem[Wolfe and Barto, 2005]{Wolfe05identifyinguseful}
Wolfe, A.~P. and Barto, A.~G. (2005).
\newblock Identifying useful subgoals in reinforcement learning by local graph
  partitioning.
\newblock In {\em ICML}.

\end{thebibliography}
\bibliographystyle{apalike}

\end{document}